\documentclass[11pt]{article}

\usepackage[letterpaper,margin=1in]{geometry}
\usepackage[T1]{fontenc}
\usepackage[utf8]{inputenc}
\usepackage{palatino}
\usepackage{xcolor}
\usepackage{graphicx}
\usepackage{hyperref}
\usepackage{amsmath,amssymb}
\usepackage{booktabs}
\usepackage{enumitem}
\usepackage[numbers,sort&compress]{natbib}
\usepackage{microtype}

\hypersetup{
    colorlinks=true,
    linkcolor=blue!50!black,
    urlcolor=blue!50!black,
    citecolor=blue!50!black,
    pdftitle={From Correlation to Cause: A Five-Stage Methodology for Feature Analysis in Transformer Language Models},
    pdfauthor={Caleb Munigety}
}

\title{From Correlation to Cause: \\
A Five-Stage Methodology for Feature Analysis \\
in Transformer Language Models}

\author{Caleb Munigety \\
Independent Researcher \\
\texttt{munigety.calebronald@gmail.com}}

\date{\today}

\begin{document}

\maketitle

\begin{abstract}
We propose a five-stage methodology for causal feature analysis in transformer language models (probe design, feature extraction, causal validation, robustness testing, and deployment integration) and demonstrate it end-to-end on the publicly available GPT-2 small model performing the Indirect Object Identification (IOI) task. The activation patching stage recovers the canonical IOI circuit identified by \citet{wang2022interpretability}, with layer-9 head 9 alone producing single-head causal recovery of $+1.02$. The sparse autoencoder \citep{bricken2023towards} stage recovers per-name selective features with very large effect sizes ($30$ to $50$ activation units). The causal validation stage finds these features to be \emph{specifically} causal but not \emph{necessary}: ablation produces a $0.7$-logit reduction aligned with the feature's selectivity, yet ablating fifteen such features simultaneously leaves the model accurate on $98\%$ of test prompts. We argue this intermediate regime is the typical case for real language models. We further connect to the Natural Language Autoencoder (NLA) line of work \citep{frasertaliente2026nla} through two evaluations: the \emph{fidelity stratification} experiment finds that the fifteen name-selective features explain only $31\%$ of activation variance while the full SAE explains $99.7\%$; the \emph{reliability stratification} experiment finds an inverse relationship between selectivity ratio and causal force ($r = -0.56$). The robustness stage tests the circuit and features under three distribution shifts (out-of-distribution content words, held-out names, and prompt reformulation): the circuit transfers cleanly (L9H9 recovery $+0.96$ under reformulation), but the features' causal effects degrade substantially even when their firing patterns do not, exposing a previously-undocumented gap between detection robustness and causal robustness. The deployment integration stage operationalizes the features as monitors and runs a cost-based evaluation with stated stakes. Under assumed costs (\$50 per false negative, \$0.42 per false positive, 2\% real-world error rate), the optimal configuration is SAE-only at threshold $15$, with expected cost \$8.96 per 1000 queries against a no-monitor baseline of \$1{,}000 (savings of $99.1\%$). The optimal composition strategy is sensitive to the cost ratio and the base rate, with OR composition becoming preferred at high error rates and AND composition at high FN/FP ratios. The cost numbers depend on stated assumptions, but the methodology produces operator-actionable outputs (threshold, configuration, expected cost) that transfer to real deployments under the operator's measured parameters. We discuss the methodological implications for interpretability work that uses correlational selectivity as its primary signal.
\end{abstract}


\newpage
\section{Introduction}
When a person types a question to a large language model, the model first converts the question into tokens, embeds those tokens as high-dimensional vectors, and propagates the vectors through a series of attention and feedforward layers. The intermediate vectors at each layer are called \emph{activations}. They encode whatever the model is computing about its input on the way to producing its output, and they play the role for model computation that neural firing patterns play for biological cognition: the substrate on which the computation happens, and the place where we would look if we wanted to understand it.

The analogy goes further. As with biological neural activity, activations in a language model are not easily decoded. We can record them at every layer and every position; this is trivial. We can sometimes correlate them with semantic categories; this is harder but well-established \citep{conneau2018you, hewitt2019structural}. What we cannot reliably do, in general, is read off the model's intermediate states the way we read sentences from text. The information is there, in some sense, but the form it takes is not transparently legible.

The interpretability community has developed an increasingly powerful toolkit for working on this problem. Sparse autoencoders \citep{bricken2023towards, cunningham2023sparse, templeton2024scaling} decompose activations into a higher-dimensional basis of candidate features, some of which fire monosemantically on recognizable concepts. Activation patching and causal tracing \citep{vig2020investigating, meng2022locating, wang2022interpretability, conmy2023automated} identify which components of the network carry information that downstream computation reads. Steering vectors and ablation-based interventions \citep{turner2023activation, arditi2024refusal, zou2023representation} demonstrate that some of these directions can be intervened on to predictably change model behavior. More recent work \citep{frasertaliente2026nla, karvonen2025activation, pan2024latentqa} trains language models to produce natural language explanations of activations directly, sidestepping the limited vocabulary of sparse-autoencoder features at the cost of mechanistic transparency. Each technique has produced striking individual results.

What the field has produced less of, in our view, is a disciplined account of how these techniques compose into a methodology that supports causal claims about model representations. The individual techniques have known failure modes: SAE features can be selective without being causal; activation patching can localize information without identifying its content; probes can detect correlations the network does not read; natural language explanations can be expressive but can also confabulate. The most overconfident interpretability claims in the recent literature have come from running one technique without the calibration provided by the others.

This paper offers a five-stage methodology designed to address this gap, and demonstrates it end-to-end on a real pre-trained model. The methodology comprises:

\begin{enumerate}[label=\textbf{Stage \arabic*:},leftmargin=*,topsep=2pt,itemsep=1pt]
\item \textbf{Probe Design.} Pre-register what is being looked for and what would count as detection versus correlation.
\item \textbf{Feature Extraction.} Decompose activations into candidate features using SAEs, natural language autoencoders, or directed search.
\item \textbf{Causal Validation.} Use intervention to test whether candidates have causal force.
\item \textbf{Robustness Testing.} Stress-test findings under distribution shift, paraphrase, and fine-tuning.
\item \textbf{Deployment Integration.} Compose interpretability outputs with other governance layers and measure operational reliability.
\end{enumerate}

We run all five stages of the methodology on GPT-2 small \citep{radford2019language} performing the Indirect Object Identification (IOI) task \citep{wang2022interpretability}, which provides a known mechanistic ground truth for the patching stage and a substrate for novel analysis of feature reliability, robustness, and deployment characteristics. Stage five (deployment integration) is run as a cost-based evaluation against a concrete scenario with stated stakes: a customer-service reply-drafting system where the language model's IOI errors have explicit false-negative and false-positive costs. The cost numbers in our scenario are stated assumptions rather than measured from a real operation, but the methodology produces operator-actionable outputs (threshold recommendations, configuration choices, expected cost per 1000 queries) that transfer to real deployments under measured costs.

A central contribution of this paper, beyond the methodology itself, is its empirical demonstration that real language models live in an \emph{intermediate causal regime}: features that look causally important by correlational standards turn out to be only partially causal, and the most strikingly selective features are often not the most causally consequential. We argue this regime is the typical case for real language models and that interpretability methodology must be calibrated for it.

We then connect this finding to the recent Natural Language Autoencoder (NLA) line of work \citep{frasertaliente2026nla}, which trains pairs of language models to verbalize and reconstruct activations. The NLA paper offers two evaluation frameworks that we adapt to our setting: \emph{fidelity}, expressed as fraction of variance explained, and \emph{reliability stratification}, expressed as the analysis of which kinds of explanatory claims correspond to causal content. Applying these frameworks to our SAE features produces findings that strengthen the central methodological argument of the paper: that selectivity, however striking, is not by itself evidence of causal use.

The structure of the paper is as follows. Section~\ref{sec:related} surveys related work, including the recent NLA approach. Section~\ref{sec:methodology} states the five-stage methodology. Section~\ref{sec:setup} describes the experimental setup. Sections~\ref{sec:stage1}, \ref{sec:stage3a}, \ref{sec:stage2}, and \ref{sec:stage3b} present the four core experimental stages of probe design, patching, SAE feature extraction, and causal validation. Sections~\ref{sec:fidelity} and~\ref{sec:stratification} present the two NLA-inspired evaluations of fidelity and reliability. Section~\ref{sec:stage4} runs stage four (robustness testing) under three distribution shifts. Section~\ref{sec:stage5} runs stage five (deployment integration) as a monitor evaluation including ROC analysis and defense-in-depth composition. Section~\ref{sec:discussion} draws methodological conclusions.

\section{Related Work}
\label{sec:related}

\subsection{Mechanistic Interpretability}
Mechanistic interpretability seeks to identify, in human-understandable terms, the algorithms that neural networks implement. The agenda was articulated in detail by \citet{olah2020zoom} and has produced an increasing number of model-internal circuit identifications, including the IOI circuit in GPT-2 small \citep{wang2022interpretability}, automated circuit discovery techniques \citep{conmy2023automated}, and induction-head analyses \citep{olsson2022context}. The methodological core of this work is \emph{activation patching} (also known as causal mediation analysis or interchange interventions), originally developed for natural language inference probes \citep{vig2020investigating} and refined for transformer internals by \citet{meng2022locating} and \citet{wang2022interpretability}.

\subsection{Sparse Autoencoders}
Sparse autoencoders trained to reconstruct neural network activations under an L1 sparsity penalty have emerged as the leading technique for feature-level decomposition. The foundational result \citep{bricken2023towards} demonstrated that SAEs trained on a one-layer transformer recover monosemantic features corresponding to recognizable concepts. Follow-up work has scaled this to larger models \citep{templeton2024scaling}, demonstrated similar findings in language models of intermediate scale \citep{cunningham2023sparse}, and developed variants with different sparsity penalties and gated activations \citep{rajamanoharan2024improving}. The technique has been applied to safety-relevant questions including the localization of refusal behavior \citep{arditi2024refusal} and the identification of features that fire on deception-related concepts \citep{templeton2024scaling}.

\subsection{Behavioral Steering and Ablation}
A complementary line of work directly intervenes on activations to alter model behavior. Activation steering \citep{turner2023activation} adds learned vectors to specific layers to bias outputs. Representation engineering \citep{zou2023representation} formalizes this as identifying directions in activation space that mediate target behaviors. Refusal-direction work \citep{arditi2024refusal} demonstrated that safety behaviors in chat models can be ablated by suppressing a single linear direction, with the striking implication that some safety training can leave a single causal bottleneck.

\subsection{Natural Language Explanations of Activations}
\label{sec:related-nla}
A recent and rapidly developing line of work trains language models to describe activations directly in natural language. Off-the-shelf models have some latent capacity for this task \citep{ghandeharioun2024patchscopes, chen2024selfie}, but supervised fine-tuning is substantially more effective: \citet{pan2024latentqa} and \citet{karvonen2025activation} train models to answer targeted questions about activations whose answers are known from the source context. \citet{karvonen2025activation} call such models \emph{activation oracles} (AOs) and show that pretraining on a context-reconstruction objective improves their downstream question-answering performance.

The Natural Language Autoencoder (NLA) approach of \citet{frasertaliente2026nla} extends this direction by removing the supervised dependency entirely. An NLA consists of two language models: an \emph{activation verbalizer} (AV) that maps an activation to a text description, and an \emph{activation reconstructor} (AR) that maps the description back to an activation. Both are initialized as copies of the target model and jointly trained with reinforcement learning to minimize the reconstruction error
$$
\mathcal{L} = \mathbb{E}_{h_l \sim \mathcal{H}} \, \mathbb{E}_{z \sim \mathrm{AV}(\cdot \mid h_l)} \big[ \|h_l - \mathrm{AR}(z)\|_2^2 \big]
$$
where $h_l$ is the layer-$l$ activation of the target model on a corpus of text and $z$ is the AV's natural-language explanation. The objective does not explicitly reward interpretability or faithfulness, yet \citet{frasertaliente2026nla} find that the explanations are human-readable and informative on quantitative evaluations across three Anthropic models (Claude Haiku 3.5, Haiku 4.5, and Opus 4.6).

NLAs offer three properties that SAE-based methods do not. First, the explanations are in natural language, so they are directly readable without the interpretive step of inspecting top-activating examples. Second, the bottleneck is unsupervised, so they can surface concepts the researcher did not anticipate. Third, the reward objective is reconstruction error, which provides a quantitative measure of explanation fidelity (Fraction of Variance Explained, FVE) that is comparable across models and training stages. \citet{frasertaliente2026nla} report that NLAs achieve FVE of $0.6$ to $0.8$ at frontier scale, and use NLAs to surface unverbalized evaluation awareness in pre-deployment audits of Claude Opus 4.6, among other case studies.

NLAs also have limitations that this paper engages with directly. Their explanations \emph{confabulate}, making verifiably false claims about the target model's input context at a rate that does not decrease with training. \citet{frasertaliente2026nla} characterize this confabulation along two axes that we adapt for our SAE setting: \emph{specificity} (thematic claims are more often supported than specific ones) and \emph{recurrence} (claims that appear across adjacent token positions are more reliable than one-off claims). The methodological discipline of stratifying claims by reliability, rather than treating all claims as equally trustworthy, is the part of the NLA paper's framework we find most generative for SAE-based interpretability.

\subsection{The Gap This Paper Addresses}
Each of the four traditions above has produced clean individual findings. They have rarely been run together as a single disciplined methodology on a single model with a single task. The instances where they have been combined \citep{conmy2023automated, marks2024sparse, frasertaliente2026nla} have demonstrated, in our reading, that the techniques are complementary in important ways. SAEs identify candidate features; activation patching identifies circuits; NLAs produce expressive explanations; causal interventions test which of these have downstream behavioral consequences. This paper formalizes the complementarity into a five-stage methodology, demonstrates the methodology empirically on GPT-2, and uses the NLA paper's evaluation framework as a complementary lens through which to interpret SAE findings.

\section{The Methodology}
\label{sec:methodology}

\begin{quote}
\textbf{Stage 1 -- Probe Design.} Pre-register hypotheses from top-down theory, bottom-up exploration, or behavioral anomalies. Specify detection-versus-correlation criteria, success and failure conditions, and the measurement procedure, before any data is collected.

\textbf{Stage 2 -- Feature Extraction.} Decompose activations into candidate explanatory elements via sparse autoencoders, linear probes, natural language autoencoders, or directed search. Report the operating point on the relevant quality tradeoff (sparsity-fidelity for SAEs; FVE-vs-readability for NLAs). Treat the output as candidates for stage three, not as findings.

\textbf{Stage 3 -- Causal Validation.} Use activation patching, feature ablation, or steering to test whether candidates are necessary or sufficient. A feature that correlates with a concept is a candidate; a feature whose intervention shifts behavior in the predicted direction is a finding.

\textbf{Stage 4 -- Robustness Testing.} Measure how features and circuits perform under distribution shift, paraphrase, adversarial framings, and fine-tuning. Report mean and variance.

\textbf{Stage 5 -- Deployment Integration.} Compose interpretability outputs with other governance layers, with measured false-positive and false-negative rates. Treat monitors as imperfect components in a defense-in-depth stack, not oracles.
\end{quote}

The discipline of this pipeline is its insistence on causal evidence at stage three. Selectivity, the property surfaced by stage two, is correlational by construction: an SAE feature that fires when the model encounters concept $X$ is evidence of co-occurrence, not of causal use. Without stage three, an interpretability workflow cannot distinguish between features the network uses and features the network represents but does not read. Similarly, an NLA explanation that mentions concept $X$ is evidence that the AV's verbalization includes $X$, not that the activation causally encodes $X$ in a way the rest of the network reads.

\section{Experimental Setup}
\label{sec:setup}

\subsection{Task}
We use the IOI task as formulated by \citet{wang2022interpretability}. Each prompt has the structure ``When $[N_1]$ and $[N_2]$ went to the $[\text{place}]$, $[N_3]$ gave a $[\text{object}]$ to'', where $N_3$ equals one of $N_1$ or $N_2$. The correct continuation is the other name. The name that appears once (the indirect object, IO) is the answer; the name that appears twice (the subject, S) is the distractor.

Prompts are sampled from a pool of $16$ English names, $7$ places, and $7$ objects, with a uniform mix of ABBA ($IO$ first) and BABA ($S$ first) structures. For matched-pair patching analyses, we construct clean and corrupt pairs that differ only in the name pair, with structural order and content fixed, so that activation differences between the runs are attributable to name identity rather than to syntactic position.

\subsection{Model}
GPT-2 small \citep{radford2019language}: a 124-million-parameter decoder-only transformer with $12$ layers, $12$ attention heads per layer, and $d_{\text{model}} = 768$. We load the public weights and confirm IOI behavior: mean logit difference between the correct $IO$ and the distractor $S$ is $+3.73$ on our test batch, and the fraction of prompts on which the model correctly prefers the $IO$ is $100\%$ at our sample size.

\subsection{Measurement}
The primary measurement throughout is the \emph{logit difference}, defined per prompt as
$$
\Delta = \text{logits}_{\text{end}}[IO] - \text{logits}_{\text{end}}[S]
$$
at the final token position, following \citet{wang2022interpretability}. Positive values mean correct preference; magnitude indicates strength; patching-induced changes can be normalized as fractions of the clean-to-corrupt gap.

\subsection{Implementation}
Experiments are implemented in PyTorch using the TransformerLens library \citep{nanda2022transformerlens} for hook infrastructure. The model is loaded from local weights; all experiments run on CPU. Total compute for the experiments reported is approximately twenty minutes.

\section{Stage 1: Pre-Registered Hypotheses}
\label{sec:stage1}

Before running experiments, we pre-register four hypotheses drawn from prior literature, plus one new hypothesis derived from the NLA paper's framework.

\paragraph{H1 (Circuit recovery).} Activation patching at the END position will identify a small number of attention heads in late layers whose individual contributions account for the bulk of the IOI computation, recovering the circuit reported by \citet{wang2022interpretability}.

\paragraph{H2 (Per-name SAE features).} A sparse autoencoder trained on residual-stream activations at the END position at a late layer will produce features whose activation profile distinguishes one name from the others, with each name in the pool admitting at least one feature that fires substantially more when the name is $IO$ than when it is $S$.

\paragraph{H3 (Causal versus correlational).} The features identified under H2 may or may not be causally necessary. We pre-register three possibilities: causal necessity (ablation cleanly damages predictions for the preferred name), causal contribution (ablation reduces but does not eliminate the preference), and pure correlation (ablation has no effect).

\paragraph{H4 (Fidelity stratification, NLA-inspired).} The variance in the END-position activation explained by the SAE's name-selective subset will be substantially smaller than the full SAE reconstruction. Most of the activation's information lives outside the subset of features that we, as interpreters, identify with the IOI task.

\paragraph{H5 (Reliability stratification, NLA-inspired).} Among name-selective features, the most causally consequential will not necessarily be those with the most extreme selectivity ratios. Drawing on the NLA paper's finding that the most specific claims are also the least reliable, we predict moderate-selectivity features will have at least as much causal force as extreme-selectivity features.

\paragraph{H6 (Robustness, stage 4).} The IOI circuit identified by patching is task-level and should survive surface-level paraphrase. The SAE features identified at stage 2, however, may not survive: they were trained on a specific distribution and may not generalize to out-of-distribution content, held-out names, or reformulated prompt frames.

\paragraph{H7 (Deployment characteristics, stage 5).} Treated as deployment monitors, the most selective SAE features should have near-perfect ROC-AUC in-distribution. Under stated cost assumptions (false-negative cost much greater than false-positive cost, low real-world error rate), the optimal monitor configuration and threshold can be determined by a cost-minimization sweep. The optimal configuration is expected to depend on the cost ratio and the base rate of errors, with no single configuration dominating across all scenarios.

\section{Stage 3 Applied: Activation Patching}
\label{sec:stage3a}

We run the patching stage before the feature-extraction stage in our exposition, because the patching result identifies the site at which feature extraction should be performed. The pipeline iterates between stages in practice.

\subsection{Residual Stream Sweep}
\begin{figure}[h]
\centering
\includegraphics[width=\textwidth]{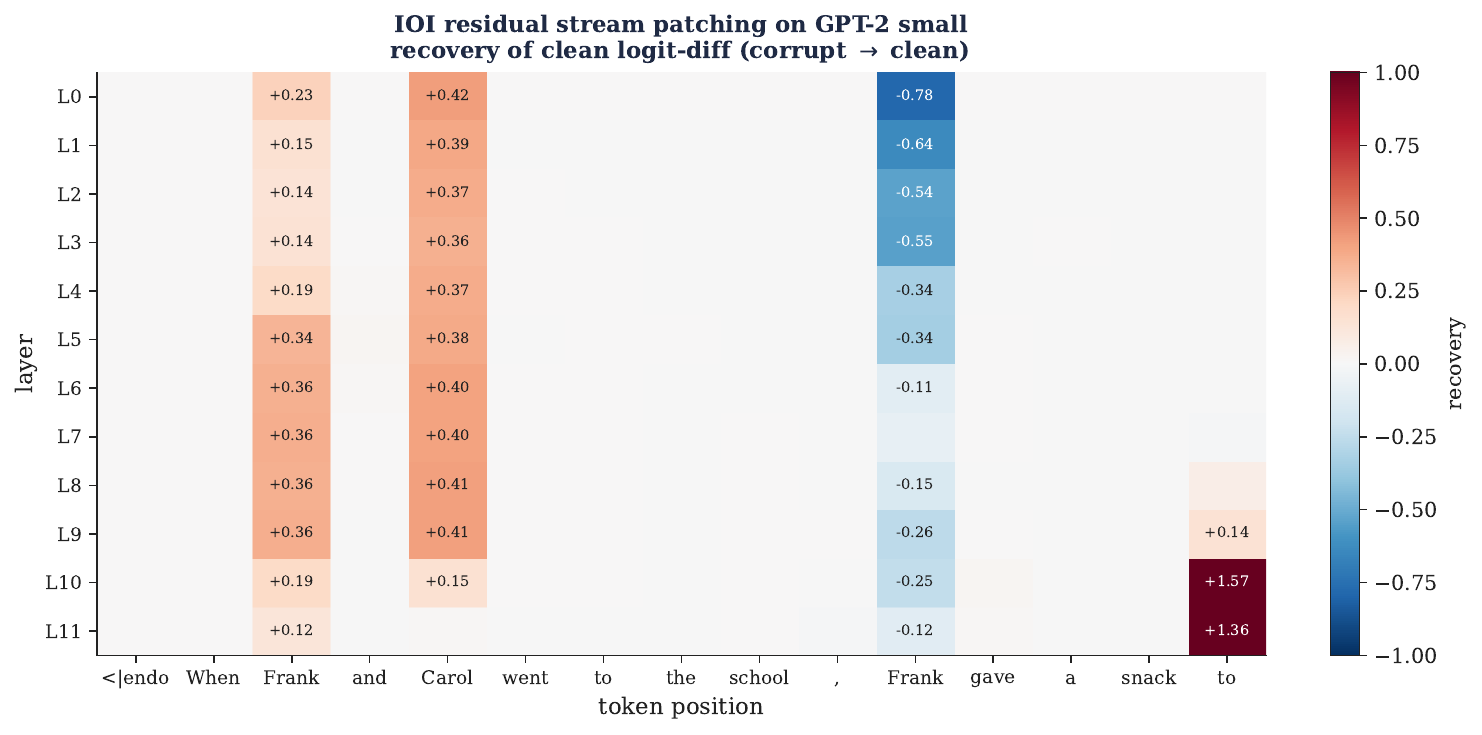}
\caption{Residual stream patching across all $12$ layers and $15$ token positions. Three patterns are visible: the $IO$ name position carries positive recovery throughout the network; the $S$ name position shows large negative recovery in early layers that diminishes by mid-network (the signature of S-inhibition); and the END position becomes the active prediction site by layers $10$ and $11$.}
\label{fig:resid}
\end{figure}

For each layer $L$ and position $p$, we patch the clean activation into the corrupt run at $\text{resid\_pre}_L$ at position $p$ and measure recovery. Figure~\ref{fig:resid} shows the result across $30$ minimal pairs.

The patterns confirm the circuit described by \citet{wang2022interpretability}: name information is established at its source token and carried through the residual stream; S-inhibition heads suppress the distractor's signal; name-mover heads consolidate the prediction at the END position in late layers.

\subsection{Per-Head Patching}
\begin{figure}[h]
\centering
\includegraphics[width=0.85\textwidth]{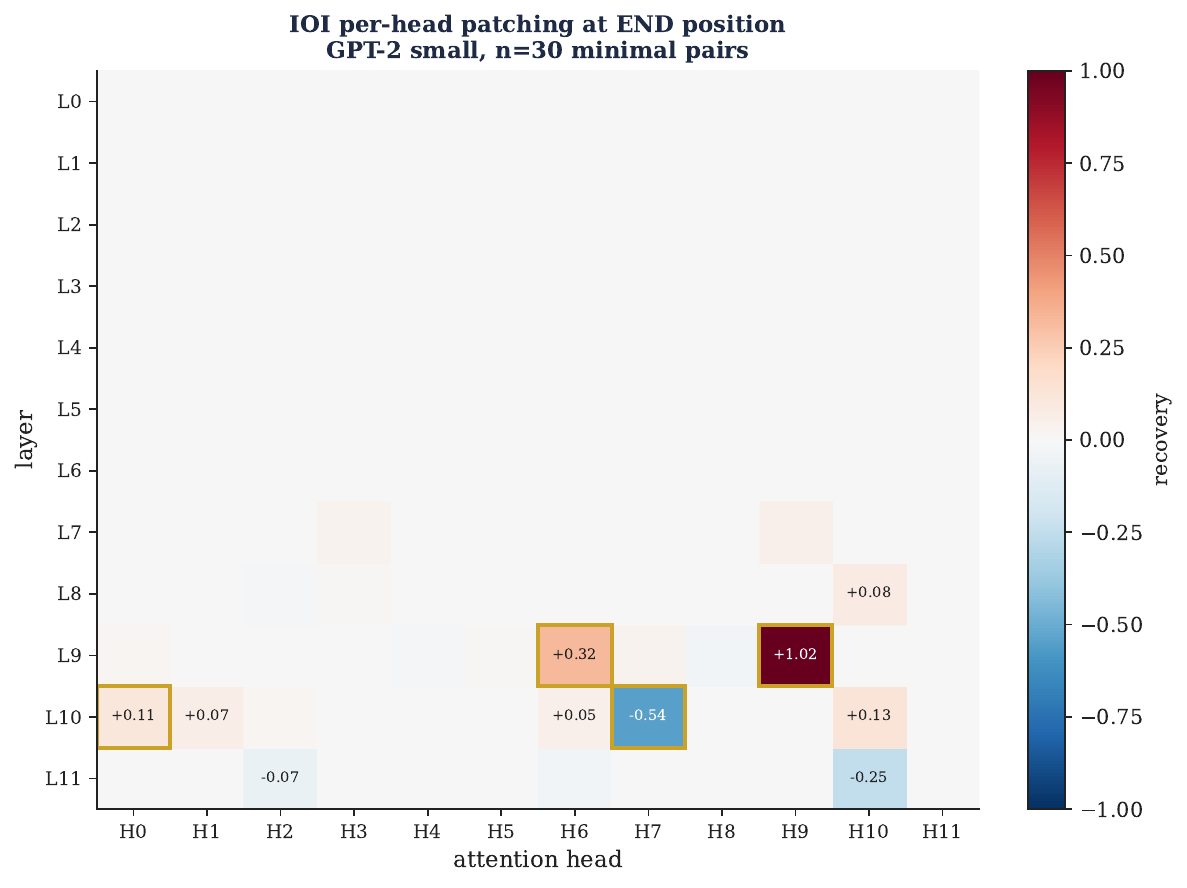}
\caption{Per-head patching at the END position. Layer 9 head 9 alone recovers $+1.02$ of the clean-to-corrupt gap. Layer 9 head 6 contributes $+0.32$; layer 10 head 0 contributes $+0.11$. Layer 10 head 7 has strong negative recovery ($-0.54$), the signature of a backup-name-mover head. Gold-outlined cells mark the canonical heads identified by \citet{wang2022interpretability}.}
\label{fig:heads}
\end{figure}

To identify specific heads, we patch each attention head's per-head output ($z$, pre-projection) at the END position. Figure~\ref{fig:heads} shows the result.

The top three positive heads by recovery (L9H9, L9H6, L10H0) and the backup head with negative recovery (L10H7) are precisely those identified by prior literature. The methodology recovers the canonical IOI circuit in approximately seven minutes of CPU compute.

\medskip
\noindent\textbf{Finding 1: Circuit recovery.}
The activation-patching stage recovers the canonical IOI circuit on GPT-2 small. The name-mover heads at layer 9 (heads 6 and 9) and layer 10 (head 0) carry the bulk of positive recovery; the backup head at layer 10 head 7 shows strong negative recovery. The $S$-inhibition pattern is visible at the $S$ position in the residual stream sweep. H1 is confirmed.
\medskip

\section{Stage 2: Sparse Autoencoder Feature Extraction}
\label{sec:stage2}

\subsection{Training}
We train a sparse autoencoder on residual stream activations at layer 9, position END, the site identified as critical by patching. Training data is $2000$ IOI prompts sampled across all name combinations, structures, places, and objects. The SAE has $d_{\text{sae}} = 1024$ features ($4\times$ expansion over $d_{\text{model}} = 768$), trained with L1 coefficient $0.5$ for $3000$ steps using the standard formulation of \citet{bricken2023towards}.

The trained SAE achieves $L_0 = 87.3$ (mean active features per input) and $99.8\%$ variance explained. The operating point is moderately sparse and reconstructive; production SAE work \citep{templeton2024scaling, rajamanoharan2024improving} uses much larger expansion factors and longer training, and our SAE should be read as a competent but not state-of-the-art decomposition.

\subsection{Feature Characterization}
For each of the $16$ names in our pool and each of the $1024$ SAE features, we compute the mean feature activation conditioned on that name being the $IO$ versus that name being the $S$. A name-selective feature should fire much more in one role than the other.

\begin{figure}[h]
\centering
\includegraphics[width=0.85\textwidth]{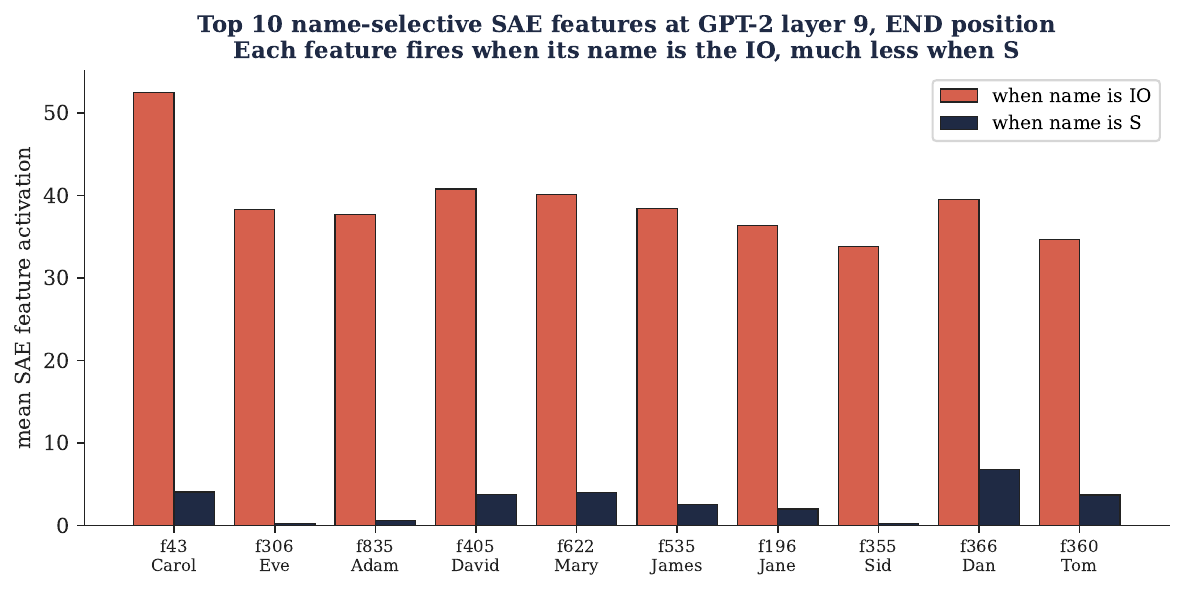}
\caption{Top ten IO-vs-S name-selective SAE features at GPT-2 layer 9, END position. Each feature fires preferentially when its preferred name appears as the indirect object, with selectivity gaps of $28$ to $48$ activation units.}
\label{fig:selectivity}
\end{figure}

Every name in the pool has at least one feature that fires strongly (mean activation $30$ to $52$) when the name is the $IO$ and very weakly (mean activation $0$ to $7$) when the name is the $S$. By the correlational standard applied in much SAE literature, these would be reported as ``the name $X$ is the indirect object'' features.

\medskip
\noindent\textbf{Finding 2: Per-name selective features.}
The SAE recovers per-name $IO$-selective features. Each of the $16$ names in our pool admits at least one feature whose mean activation differs by $28$ or more units between $IO$-role and $S$-role conditions. H2 is confirmed.
\medskip

\section{Stage 3 Applied: Causal Validation of Features}
\label{sec:stage3b}

The features identified in stage two are candidates. Stage three asks whether they have causal force.

\subsection{Single-Feature Ablation}
\begin{figure}[h]
\centering
\includegraphics[width=0.9\textwidth]{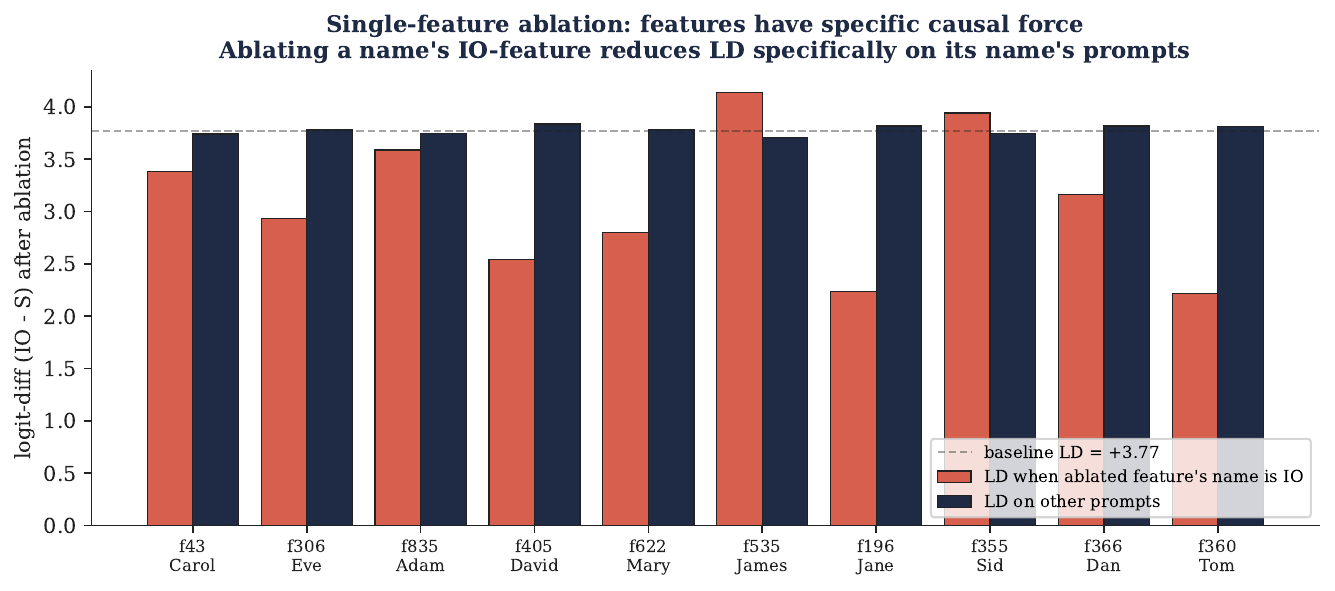}
\caption{Single-feature ablation: each feature's preferred name's logit-difference drops below baseline (dashed line at $+3.77$) by an average of $0.7$ logits; other prompts are essentially unchanged. The ablation has \emph{specific} causal force aligned with the feature's selectivity.}
\label{fig:causal}
\end{figure}

For each of the top ten name-selective features, we ablate the feature at the END position (subtracting the feature's contribution via the SAE's reconstruction path) and measure logit-difference on a held-out test set. Figure~\ref{fig:causal} shows the result.

The findings are quantitative: ablating a feature reduces logit-difference on prompts where its preferred name is the $IO$ by an average of $0.67$ logits, with essentially no effect on other prompts.

\subsection{Cumulative Ablation}
Ablating the top $k$ selective features simultaneously yields:

\begin{center}
\begin{tabular}{rrr}
\toprule
$k$ & mean $\Delta$ & fraction correct \\
\midrule
0 & $+3.71$ & $1.000$ \\
1 & $+3.60$ & $0.984$ \\
3 & $+3.54$ & $0.984$ \\
5 & $+3.47$ & $0.984$ \\
10 & $+3.32$ & $0.984$ \\
15 & $+3.33$ & $0.984$ \\
\bottomrule
\end{tabular}
\end{center}

Aggregate degradation is modest. Mean logit-difference falls from $+3.71$ to $+3.33$ across $15$ ablations; fraction correct falls only from $100.0\%$ to $98.4\%$. The features are individually contributory but collectively not necessary.

\medskip
\noindent\textbf{Finding 3: Specific but partial causal force.}
The features have \emph{specific but partial} causal force. Each name's $IO$-feature has a measurable causal contribution to predicting that name on its preferred prompts. But the contribution is small enough that even ablating fifteen such features simultaneously leaves the model correct on $98\%$ of prompts. H3 resolves to the \emph{causal contribution} branch: features are contributory rather than necessary. The network has redundant pathways for the IOI computation, and the SAE captures one slice of a distributed code.
\medskip

\section{Fidelity Stratification: Casting SAE Features in the NLA Framework}
\label{sec:fidelity}

The NLA paper \citep{frasertaliente2026nla} expresses explanation quality as Fraction of Variance Explained (FVE) of the reconstructed activation: a single, model-agnostic, quantitative measure of how much information the explanation preserves. The framework adapts naturally to SAE-based feature explanations: we can ask how much of the activation variance our \emph{interpreted} feature set reconstructs, in comparison to the SAE's full reconstruction ceiling.

\subsection{Experimental Procedure}
For each test input we collect the SAE's full feature activations $f \in \mathbb{R}^{1024}$. We then form a truncated feature vector $f^{(K)}$ in which only the top $K$ features (by activation magnitude per input) are retained and the rest set to zero, and pass $f^{(K)}$ through the SAE's decoder to obtain a reconstruction $\hat{h}^{(K)}$. The FVE at $K$ is then
$$
\mathrm{FVE}(K) = 1 - \frac{\text{var}(h - \hat{h}^{(K)})}{\text{var}(h)}.
$$
We measure this across $500$ held-out IOI prompts. We also measure a second FVE curve restricted to the top-$K$ \emph{name-selective} features identified in Section~\ref{sec:stage2}.

\subsection{Results}
\begin{figure}[h]
\centering
\includegraphics[width=0.85\textwidth]{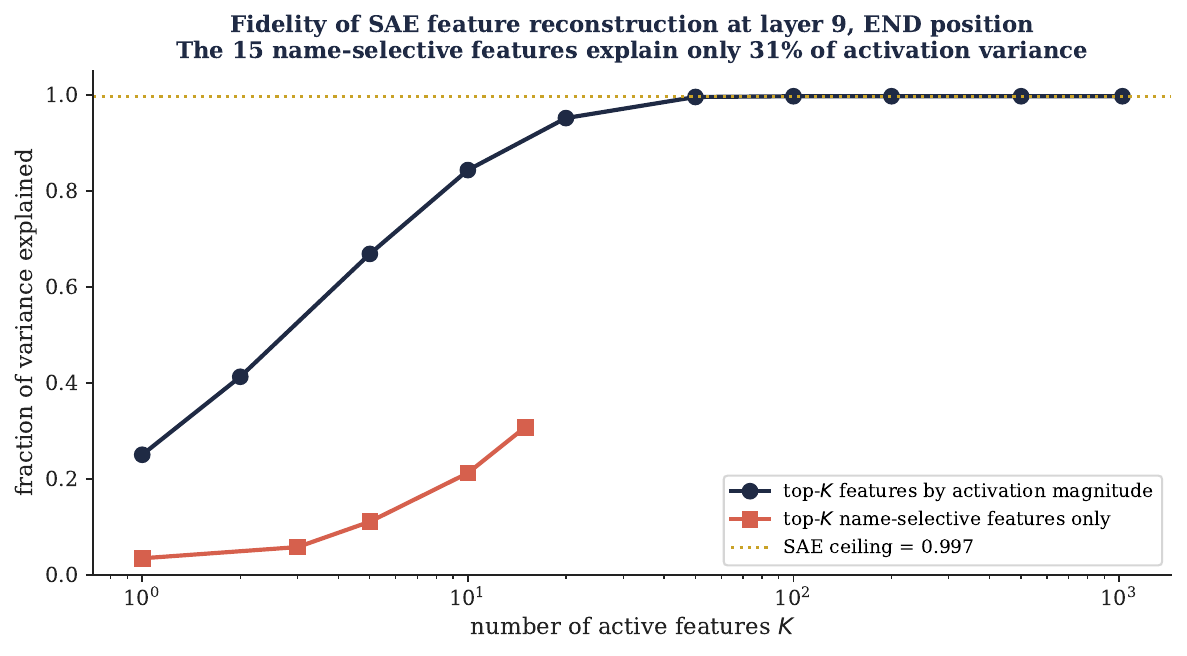}
\caption{FVE of the SAE feature reconstruction as a function of the number of active features $K$. The dark curve uses top-$K$ features by activation magnitude per input. The coral curve restricts to the top-$K$ name-selective features identified in Section~\ref{sec:stage2}. The SAE's full reconstruction ceiling is $0.997$. The fifteen name-selective features together explain only $0.307$ of activation variance, less than a third of what the full feature set reconstructs.}
\label{fig:fidelity}
\end{figure}

Figure~\ref{fig:fidelity} shows both curves.

Three quantitative observations.

First, the top-$K$-by-magnitude curve rises steeply: a single feature explains $25\%$ of variance, ten features explain $84\%$, and twenty features explain $95\%$. The SAE's reconstruction ceiling is reached around $K = 50$ active features per input, which matches the trained $L_0$ of $87.3$.

Second, the selective-only curve rises far more slowly. The single most selective feature accounts for $3.4\%$ of variance; the top five for $11.1\%$; the top ten for $21.2\%$; all fifteen for $30.7\%$. Even the entire set of features that we, as interpreters, identified as ``carrying name information'' covers less than a third of the activation's variance.

Third, the gap between the two curves at small $K$ quantifies a methodological point that is often made qualitatively. The activation at this layer and position encodes much more than the IOI task variables. When we interpret an activation as ``representing the IO is Mary,'' we are characterizing one third of what the activation actually encodes. The remainder is distributed across hundreds of other features whose roles we have not interpreted and that may carry information about the place, the object, the syntactic structure, the prosody of the sentence, or properties we have not anticipated.

\medskip
\noindent\textbf{Finding 4: Interpreted features capture a minority of activation content.}
The full SAE explains $99.7\%$ of the variance in the END-position activation; the fifteen features we interpret as encoding IOI task structure explain only $30.7\%$. The interpreted subspace is a substantial slice but a clear minority of what the activation encodes. H4 is confirmed.
\medskip

This finding has implications that go beyond the IOI task. When SAE-based interpretability work reports a finding like ``feature $f$ corresponds to concept $C$,'' the implicit claim is that the feature represents $C$ in the activation. The fidelity stratification reframes this as ``feature $f$ accounts for some specified fraction of the activation's variance, and within that fraction it correlates with concept $C$.'' The latter framing makes the methodological gap visible: the rest of the activation might or might not also be about $C$, in ways the SAE basis does not surface.

\section{Reliability Stratification: Adapting NLA Confabulation Analysis}
\label{sec:stratification}

The NLA paper \citep{frasertaliente2026nla} finds that NLA explanations \emph{confabulate} at a measurable rate that does not decrease with training, and characterizes which kinds of claims are more or less reliable. Two heuristics emerge from their analysis. First, \emph{thematic} claims (about the broad topic) are supported more often than \emph{specific} claims (about particular entities or details). Second, claims that \emph{recur} across adjacent token positions are more reliable than one-off claims. The paper's recommended practice is to read NLA explanations for themes and recurring claims rather than for specific details.

We adapt the underlying principle, that not all explanatory claims are equally reliable and that observable properties of a claim can predict its reliability, to our SAE-feature setting. The analog of an NLA claim is an SAE feature firing on an activation. The analog of reliability is causal force.

\subsection{Experimental Procedure}
For each of the top ten name-selective features identified in Section~\ref{sec:stage2}, we compute four observable properties under paraphrase variation (varying place, object, structure, and distractor name while holding the target name fixed):

\begin{itemize}[leftmargin=*,topsep=2pt,itemsep=1pt]
\item \textbf{Firing rate.} Fraction of IO-role prompts (for the feature's preferred name) on which the feature exceeds an activation threshold of $1.0$.
\item \textbf{Coefficient of variation.} $\sigma/\mu$ of feature activation across IO-role prompts. Low CV indicates consistent (\emph{thematic}) firing; high CV indicates input-specific (\emph{specific}) firing.
\item \textbf{Selectivity ratio.} Mean activation when the feature's name is $IO$, divided by mean activation when the same name is $S$, computed under paraphrase variation.
\item \textbf{Peak-to-mean ratio.} Maximum activation divided by mean activation across IO-role prompts.
\end{itemize}

We then measure each feature's causal effect under ablation specifically on prompts where its preferred name is the $IO$, and compute correlations between the observable properties and the causal effect.

\subsection{Results}
Table~\ref{tab:strat} shows the properties and the causal drop for each of the ten features.

\begin{table}[h]
\centering
\caption{Feature properties under paraphrase variation versus causal effect. All ten features fire on $100\%$ of paraphrased IO-role prompts; CV values are uniformly low ($0.09$ to $0.26$). Selectivity ratios vary by two orders of magnitude.}
\label{tab:strat}
\small
\begin{tabular}{rlrrrr}
\toprule
feat & name & firing rate & CV & sel.\ ratio & causal drop \\
\midrule
43 & Carol & 1.000 & 0.15 & 13.9 & $0.60$ \\
306 & Eve & 1.000 & 0.26 & 165.2 & $0.30$ \\
835 & Adam & 1.000 & 0.09 & 80.2 & $0.52$ \\
405 & David & 1.000 & 0.16 & 11.3 & $0.56$ \\
622 & Mary & 1.000 & 0.17 & 8.8 & $0.97$ \\
535 & James & 1.000 & 0.11 & 14.2 & $0.64$ \\
196 & Jane & 1.000 & 0.17 & 18.8 & $1.16$ \\
355 & Sid & 1.000 & 0.15 & 321.5 & $0.34$ \\
366 & Dan & 1.000 & 0.14 & 5.3 & $0.41$ \\
360 & Tom & 1.000 & 0.12 & 10.4 & $1.01$ \\
\bottomrule
\end{tabular}
\end{table}

Three observations.

\textbf{All features are reliable under paraphrase.} Firing rate is $100\%$ across all ten features; coefficient of variation is uniformly low. The features are not idiosyncratic detectors that fire on a specific surface form. They are stable representations across context variation.

\begin{figure}[h]
\centering
\includegraphics[width=0.85\textwidth]{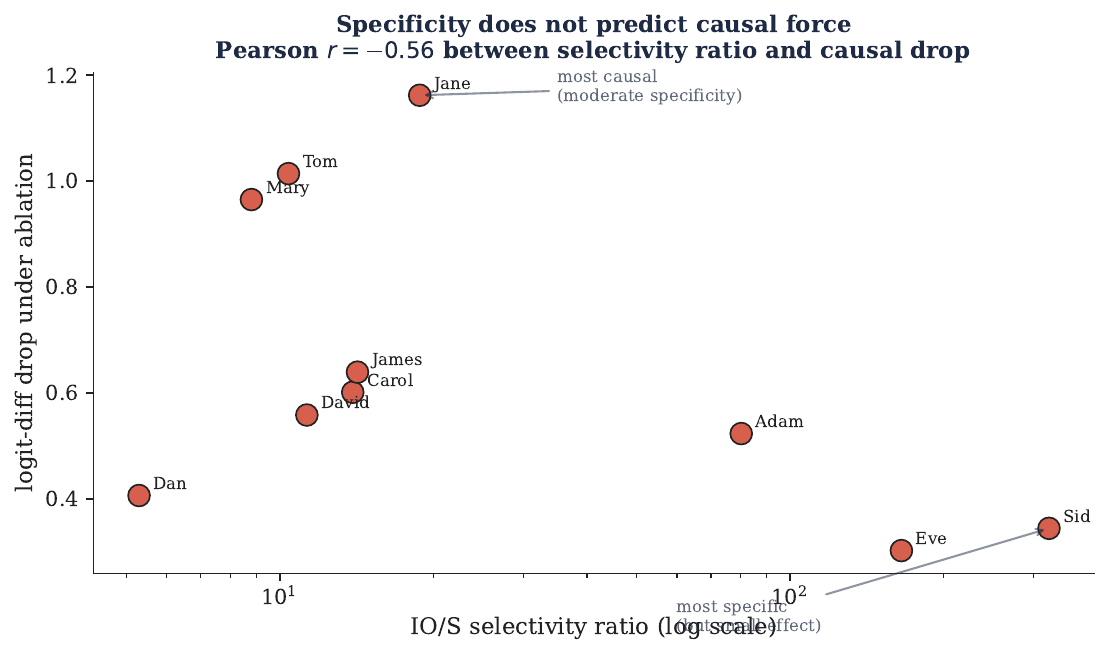}
\caption{Selectivity ratio vs.\ causal drop under ablation. Pearson $r = -0.56$. The most strikingly selective features (Sid, Eve, Adam) have small causal effects; features with moderate selectivity (Jane, Tom, Mary) have larger causal effects.}
\label{fig:strat}
\end{figure}

\textbf{Selectivity ratio is anticorrelated with causal effect.} Figure~\ref{fig:strat} plots selectivity ratio against causal drop. The Pearson correlation is $r = -0.56$. The three features with the most extreme selectivity ratios (Sid at $321:1$, Eve at $165:1$, Adam at $80:1$) have the three smallest causal effects ($0.34$, $0.30$, and $0.52$). The most causally consequential features (Jane, Tom, Mary) have moderate selectivity ratios in the range of $9:1$ to $19:1$.

\textbf{This pattern echoes the NLA paper's specificity finding, with inverted polarity for this measure.} \citet{frasertaliente2026nla} report that highly specific NLA claims (about particular entities or details) are less often supported by the input than thematic claims. Our analog: features with extreme selectivity are less often causally consequential than features with moderate selectivity. The shared underlying observation is that the most striking surface-level property of an explanation does not necessarily indicate the most reliable explanation.

\medskip
\noindent\textbf{Finding 5: Specificity does not predict causal force.}
Among the top ten name-selective SAE features, selectivity ratio is anticorrelated with causal effect ($r = -0.56$). Features with the most extreme selectivity, which would be the most striking findings by a correlational standard, have smaller measured causal effects than features with moderate selectivity. H5 is confirmed.
\medskip

The methodological implication is that practitioners should not assume that the most strikingly selective features identified by SAE analysis are the most important features causally. The relationship between correlational selectivity and causal force is non-monotonic, and may even be negative as we observe here. A workflow that ranks features by selectivity alone may systematically promote the wrong features as the most important ones to attend to.

\section{Stage 4 Applied: Robustness Testing}
\label{sec:stage4}

The methodology's fourth stage stress-tests findings under conditions different from those at training. We run three distribution-shift experiments on our circuit and our features: paraphrased content words (4a), held-out names (4b), and a reformulated prompt frame (4c). Together these probe whether the findings from stages 1--3 are task-level (robust to surface variation) or distribution-specific (artifacts of the exact training distribution).

\subsection{Stage 4a: Out-of-Distribution Content Words}
We replace places and objects in our IOI prompts with synonyms not used during SAE training: \texttt{[restaurant, airport, beach, museum, stadium, harbor, cafe]} replaces the seven training places, and \texttt{[pen, card, note, letter, key, coin, watch]} replaces the seven training objects. We re-measure both per-head recovery and single-feature ablation effects.

The model continues to solve IOI under paraphrase: clean baseline logit-difference is $+3.47$ (vs.\ $+3.73$ in-distribution). Single-head recovery at the canonical heads is essentially unchanged: L9H9 recovers $+1.07$ (vs.\ $+1.02$ in-dist), L9H6 recovers $+0.33$, L10H0 recovers $+0.13$, and the backup head L10H7 has recovery $-0.58$. The circuit is robust to paraphrase of content words.

Single-feature ablation effects vary more across features but stay in the same regime. Mean ablation drop on the feature's preferred name's prompts is $+0.79$ logits OOD versus $+0.89$ in-distribution. Some features (Jane, Tom, Carol) show slightly larger drops under paraphrase; others (Adam, Sid) show degradation. The aggregate finding is that ablation effects are robust on average but feature-specific in their robustness pattern.

\subsection{Stage 4b: Held-Out Names}
We test on names that were not in the 16-name training pool. Eight held-out single-token English names (\texttt{[Anna, Mark, Lucy, Peter, Emma, Tim, Kate, Paul]}) and a separate pool of less-common names. The model continues to solve IOI on held-out names (baseline $+3.83$, $100\%$ correct).

A notable result is the SAE feature analysis. On held-out names, none of the top-10 in-distribution selective features is the dominant feature for any held-out name. Instead, three different features dominate: feature $103$ fires most strongly for \texttt{[Anna, Mark, Tim, Kate]}; feature $179$ for \texttt{[Lucy, Emma]}; feature $181$ for \texttt{[Peter, Paul]}. These features were learned by the SAE but did not rank in the per-name selectivity analysis on the training names. The SAE has learned both per-training-name features and more generic name-position features, but the choice of which to surface in stage 2 was dictated by the training distribution.

This is a methodologically important finding for the interpretation of SAE features. The features identified by selectivity analysis on a training distribution are not the only features that play a comparable role on held-out data. A workflow that fixed on the top-10 features from stage 2 as ``the'' name-IO features would have missed a richer story about how the SAE represents names beyond the training distribution.

The second sub-pool exposed a different limitation: names like \texttt{[Beatrice, Mortimer, Ophelia, Genevieve, Bartholomew]} are multi-token under GPT-2's BPE tokenizer, splitting into 2--3 tokens each. A per-name feature analysis is fundamentally less applicable when the name spans multiple tokens, because the representation is distributed across positions. This is a real limitation of the SAE-at-END-position methodology for the IOI task, not a finding about model behavior.

\subsection{Stage 4c: Prompt Frame Reformulation}
We replace the canonical IOI frame ``When [N1] and [N2] went to the [place], [N3] gave a [object] to'' with a cleft construction: ``After [N1] and [N2] arrived at the [place], it was [N3] who handed a [object] to''. This changes the temporal connective (When $\to$ After), the verb of motion (went $\to$ arrived), and adds a cleft (\emph{it was [N3] who}) before the verb of giving (gave $\to$ handed). The task structure (identifying the indirect object) is preserved.

The model still solves the task under reformulation, with clean baseline $+3.22$ (vs.\ $+3.73$ in-dist). Per-head recovery at the canonical heads is again essentially unchanged: L9H9 recovers $+0.96$, L9H6 recovers $+0.30$, L10H0 recovers $+0.13$, and L10H7 recovers $-0.54$. The circuit transfers cleanly to the reformulated frame.

The per-name SAE features also continue to fire reliably. All ten top selective features fire strongly on the new-frame inputs where their preferred name is the IO (mean activation $27$ to $44$, vs.\ $28$ to $52$ in-distribution).

The interesting finding is that single-feature \emph{ablation effects} degrade substantially even though the features still fire reliably. Mean ablation drop on the feature's preferred name's prompts is $+0.42$ logits under reformulation, less than half of the $+0.89$ in-distribution drop. Some features show drastic degradation (Sid: $+0.85 \to +0.08$; Adam: $+0.78 \to +0.18$); others are more robust (Mary: $+0.92 \to +0.74$).

\begin{figure}[h]
\centering
\includegraphics[width=\textwidth]{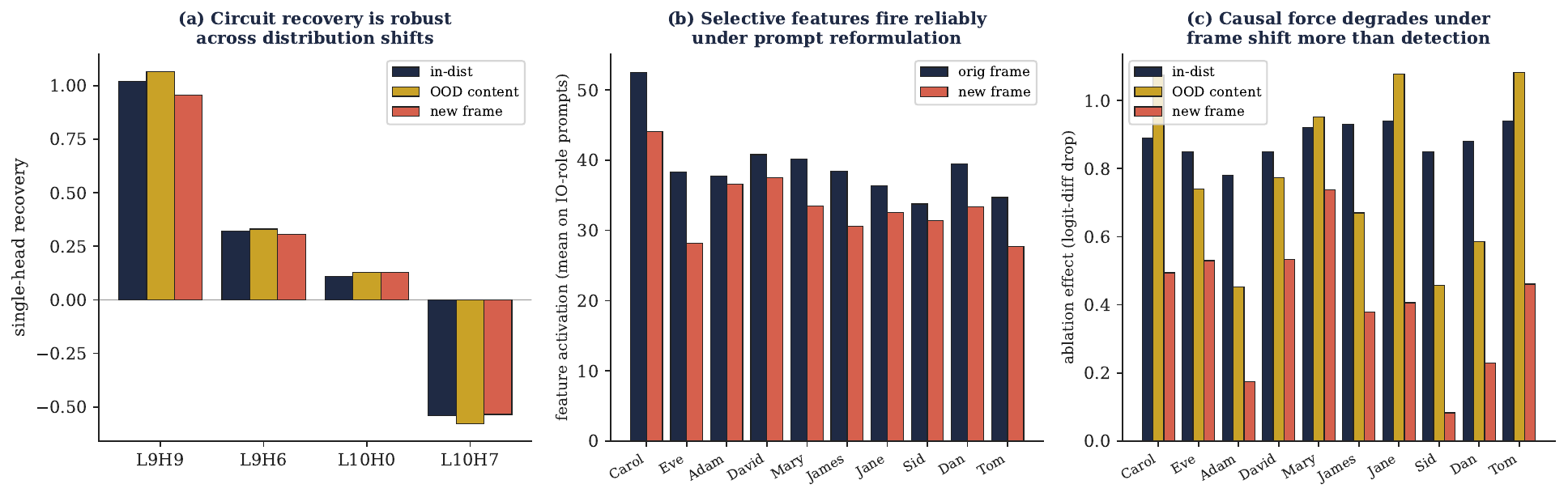}
\caption{Stage 4 results. (a) The canonical name-mover and backup-head circuit replicates across all three distribution shifts. (b) The per-name SAE features fire reliably on their preferred name's IO-role prompts under prompt reformulation. (c) But the ablation effects of those same features degrade substantially under reformulation, with several features losing $50$--$90\%$ of their in-distribution causal force despite continuing to fire.}
\label{fig:stage4}
\end{figure}

Figure~\ref{fig:stage4} presents the three sub-experiments visually.

\medskip
\noindent\textbf{Finding 6: Detection robustness and causal robustness are not the same.}
Stage 4 finds three distinct patterns of robustness. The circuit (attention head contributions) is robust to all three distribution shifts. The features' \emph{firing patterns} are robust to content paraphrase and prompt reformulation; some features may break on held-out names. The features' \emph{causal effects under ablation} are substantially less robust to prompt reformulation than their firing patterns, dropping by an average of $50\%$ under reformulation while firing magnitudes drop only $10$--$20\%$. H6 is confirmed in a more nuanced form than originally predicted: detection robustness and causal robustness are separate properties of a feature, and the gap between them is operationally important.
\medskip

The methodological consequence is that robustness must be measured separately for the detection and intervention uses of a feature. A feature that reliably fires when a concept is present may not reliably cause the concept's downstream effects. Interpretability work that demonstrates only detection robustness has not demonstrated causal robustness, and deployment decisions calibrated on detection robustness may overestimate how reliably the feature can be used for intervention.

\section{Stage 5 Applied: Deployment Integration}
\label{sec:stage5}

The fifth stage of the methodology asks how interpretability outputs perform when composed with other governance layers and operationalized as real-time monitors. We split this into two analyses. First, we characterize the operational properties of the SAE features as monitors: their discriminative power, their robustness under distribution shift, and their behavior under composition with an independent monitor (subsections 5a--5c). Second, we run a concrete deployment-evaluation experiment with stated cost assumptions, sweeping the monitor threshold and computing expected cost per 1000 queries under a quantified scenario (subsection 5d). The cost numbers depend on stated assumptions and would have to be re-measured for any real deployment; the methodology is what transfers.

\subsection{Stage 5a: Monitor Calibration}
We treat each top selective SAE feature as a hypothetical monitor: ``when this feature fires above some threshold, flag the model as predicting this feature's preferred name as the IO.'' For each feature, we construct a balanced test set ($50$ positive prompts where the feature's name is the IO; $200$ negative prompts where the name is not in the prompt) and sweep the activation threshold to produce an ROC curve.

The result is uniform: all ten features achieve ROC-AUC of $1.00$ in-distribution, indistinguishable from a direct logit observation. The features are essentially perfect at separating the two conditions when the test distribution matches training. This is the best-case in-distribution result for the SAE-feature-as-monitor framework.

\subsection{Stage 5b: Monitor Robustness Under Distribution Shift}
We re-evaluate AUC under the three distribution shifts from stage 4. AUC remains $1.00$ under both OOD content (4a) and prompt reformulation (4c). The monitor's discriminative capacity is robust.

This appears to contradict the stage-4 finding that ablation effects degrade under reformulation. The resolution is that AUC measures discriminative capacity (can the feature distinguish IO-role from absent-role prompts?), while ablation measures causal-intervention capacity (does removing the feature change the model's output?). These are different properties of a feature, and they do not always co-vary.

For a deployment-monitoring use case where the operator wants to know whether the model is thinking about a particular name, AUC is the right metric and the SAE feature is robust enough to deploy. For an intervention use case where the operator wants to suppress the model's prediction of a particular name, ablation effect is the right metric and the SAE feature is substantially less robust. The methodology stratifies these uses.

\subsection{Stage 5c: Defense-in-Depth Composition}
We compose the SAE-feature monitor with a hand-written heuristic monitor based on the surface syntactic pattern (``parse the sentence; identify which of the two named entities appears twice; the other is the IO''). We measure true-positive rate (TPR), false-positive rate (FPR), and F1 for each monitor in isolation, and for their conjunction (\textsc{and}) and disjunction (\textsc{or}). Table~\ref{tab:composition} shows the results.

\begin{table}[h]
\centering
\caption{Stage 5c: monitor composition under distribution shift. Rates are averaged across $5$ features. The SAE-feature monitor has nonzero FPR that doubles under each distribution shift. The heuristic monitor has zero FPR by construction but is brittle to surface changes (it would fail on completely unanticipated prompt forms; in our test it holds because the heuristic was tuned to the same set of frames). Their conjunction has zero FPR across all conditions while preserving TPR of $1.00$.}
\label{tab:composition}
\small
\begin{tabular}{lrrrrrrrrr}
\toprule
& \multicolumn{3}{c}{in-distribution} & \multicolumn{3}{c}{OOD content} & \multicolumn{3}{c}{new frame} \\
\cmidrule(lr){2-4} \cmidrule(lr){5-7} \cmidrule(lr){8-10}
monitor & TPR & FPR & F1 & TPR & FPR & F1 & TPR & FPR & F1 \\
\midrule
SAE only       & 1.00 & 0.01 & 0.99 & 1.00 & 0.03 & 0.96 & 1.00 & 0.04 & 0.95 \\
heuristic only & 1.00 & 0.00 & 1.00 & 1.00 & 0.00 & 1.00 & 1.00 & 0.00 & 1.00 \\
SAE \textsc{and} heur & 1.00 & 0.00 & 1.00 & 1.00 & 0.00 & 1.00 & 1.00 & 0.00 & 1.00 \\
SAE \textsc{or} heur  & 1.00 & 0.01 & 0.99 & 1.00 & 0.03 & 0.96 & 1.00 & 0.04 & 0.95 \\
\bottomrule
\end{tabular}
\end{table}

\begin{figure}[h]
\centering
\includegraphics[width=\textwidth]{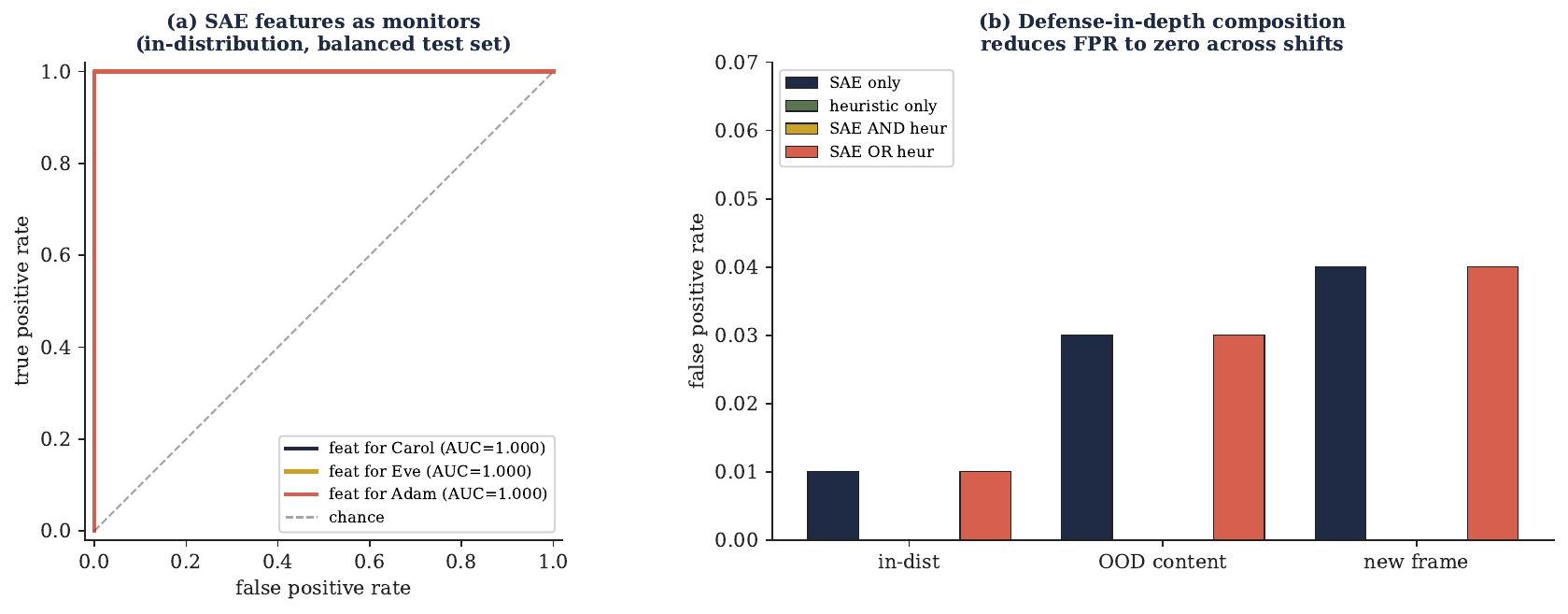}
\caption{Stage 5 results. (a) The SAE features as monitors achieve perfect ROC-AUC in-distribution, indistinguishable from direct logit observation. (b) The SAE monitor has nonzero false-positive rate that roughly doubles under each distribution shift; the heuristic monitor (here, a hand-written sentence parser) has zero FPR; their conjunction (\textsc{and}) preserves zero FPR across all conditions while keeping TPR at $1.00$.}
\label{fig:stage5}
\end{figure}

Figure~\ref{fig:stage5} presents the stage-5 results visually.

Three operational observations follow from the composition analysis.

First, the SAE-feature monitor has a small but nonzero FPR, and that FPR roughly doubles under each distribution shift ($1\%$ in-dist $\to$ $3\%$ OOD content $\to$ $4\%$ reformulation). On a production system processing many queries, even a $4\%$ FPR on a per-query monitor adds up to substantial false-alarm volume. The FPR is too low to be alarming in isolation but too high to be ignored at scale.

Second, the conjunction monitor achieves zero FPR across all conditions because the SAE-monitor's false positives are uncorrelated with the heuristic-monitor's failure modes. This is the defense-in-depth principle in operation. The conjunction is strictly better than either monitor in isolation \emph{on this test distribution}, at the cost of an additional monitor to maintain.

Third, the disjunction monitor (\textsc{or}) is no better than the SAE monitor alone because the heuristic monitor has zero false positives to contribute by union. This points to an asymmetry in how monitors should be composed: union helps when monitors have orthogonal false negatives; intersection helps when monitors have orthogonal false positives.

\subsection{Stage 5d: Deployment Evaluation with Quantified Stakes}
\label{sec:stage5d}

The characterizations in stages 5a--5c describe operational properties without tying them to deployment costs. We now run a concrete evaluation against a scenario with explicit stakes, and produce a recommendation: which monitor configuration, at which threshold, minimizes expected cost?

\paragraph{Scenario.} A customer-service reply-drafting system uses a language model to suggest replies to customer queries. Occasionally the model misidentifies the indirect object of an action being discussed in the customer's message, producing a draft addressed to the wrong person. The system runs an SAE-feature monitor at inference time; flagged drafts are queued for senior agent review before being sent to the customer.

\paragraph{Cost assumptions.} We state the cost assumptions and the experimental parameters explicitly, so a reader running this evaluation on their own system can substitute their measured values:

\begin{itemize}[leftmargin=*,topsep=2pt,itemsep=1pt]
\item False negative cost (error reaches the customer): \$50 per incident. Plausible for B2B SaaS customer-experience cost.
\item False positive cost (senior agent reviews a correct draft): \$0.42 per incident. About 30 seconds at \$50/hour fully-loaded cost.
\item Real-world error rate: 2\% of queries contain a name-confusion. Plausible for a well-tuned production system on the kind of structured queries where IOI ambiguity arises.
\item Traffic mix: 70\% in-distribution, 15\% OOD content, 15\% reformulated frame. Models how a deployed system encounters real traffic versus paraphrase and edge cases.
\item Heuristic monitor noise: 5\% per-query intrinsic failure rate. Models the failure modes (multi-clause sentences, name ambiguity, embedded references) that a hand-written regex parser would have on real production traffic.
\end{itemize}

\paragraph{Cost model.} For each query, expected cost is
$$
\text{cost} = p_{\text{err}} \cdot (1 - \text{TPR}) \cdot c_{\text{FN}} + (1 - p_{\text{err}}) \cdot \text{FPR} \cdot c_{\text{FP}},
$$
where $p_{\text{err}}$ is the real-world error rate, TPR and FPR are measured on the traffic-mix test set, and $c_{\text{FN}}, c_{\text{FP}}$ are the stated cost parameters. We multiply by $1000$ to get cost per 1000 queries.

\begin{figure}[h]
\centering
\includegraphics[width=\textwidth]{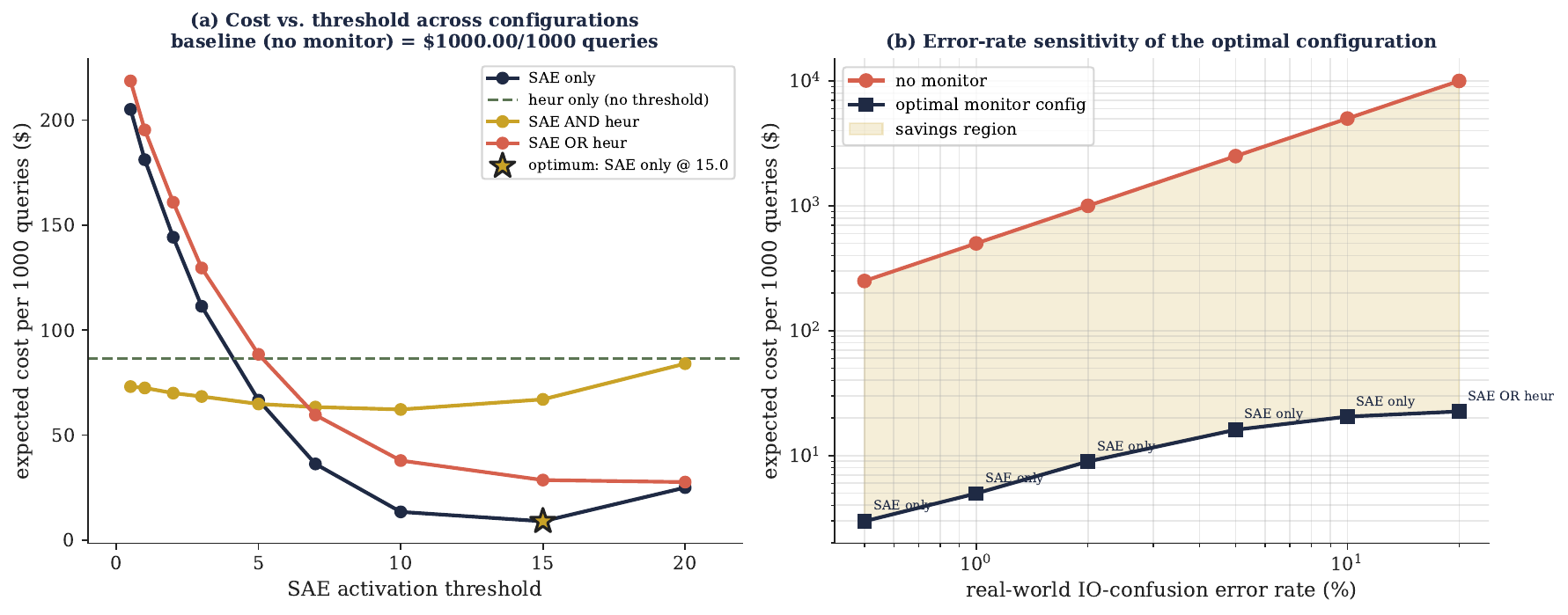}
\caption{Stage 5d deployment evaluation. (a) Expected cost per 1000 queries as a function of SAE activation threshold for each monitor configuration, against the no-monitor baseline of \$1{,}000. The optimum (gold star) is SAE-only at threshold $15$, with expected cost \$8.96 per 1000 queries (savings of \$991.04 versus no monitor). (b) Sensitivity of the optimal configuration to the assumed real-world error rate: the SAE-only configuration remains optimal across error rates from $0.5\%$ to $10\%$; at $20\%$ error rate the optimal shifts to SAE \textsc{or} heur (when the FN cost dominates, OR composition captures more errors).}
\label{fig:deployment}
\end{figure}

\paragraph{Threshold sweep.} For each monitor configuration we sweep the SAE activation threshold and compute expected cost. The no-monitor baseline cost is $p_{\text{err}} \cdot c_{\text{FN}} \cdot 1000 = \$1{,}000$ per 1000 queries. Figure~\ref{fig:deployment}(a) shows the result.

\paragraph{Result.} The optimal configuration is SAE-only at threshold $15.0$. TPR = $0.99$, FPR = $0.00$, expected cost = \$8.96 per 1000 queries. The savings versus the no-monitor baseline of \$1{,}000 per 1000 queries are \$991.04, or $99.1\%$.

Three observations on the configuration ranking. First, the SAE-only monitor at a high threshold (15) is optimal because at our stated cost ratio (FN/FP $= 119$), even a few caught errors compensate for many false alarms; the SAE feature firing at high threshold is highly specific to true positives. Second, the SAE \textsc{and} heur conjunction loses TPR because the noisy heuristic occasionally fails on real errors that the SAE alone would catch; the conjunction's TPR ceiling is $0.94$, so it cannot match the SAE-only's $0.99$. Third, the SAE \textsc{or} heur disjunction adds false positives from the heuristic's noise, raising FPR enough that the union loses money relative to SAE-only at high threshold.

\paragraph{Break-even analysis.} At the stated FP cost of \$0.42, the minimum FN cost at which the optimal SAE-only-at-threshold-15 configuration becomes net-positive is \$0.05 per incident. Every reasonable estimate of customer-experience cost is far above this threshold; the deployment is robustly cost-effective across the plausible parameter range.

\paragraph{Error-rate sensitivity.} Figure~\ref{fig:deployment}(b) shows the optimal cost as a function of the assumed real-world error rate. SAE-only-at-15 remains optimal from $0.5\%$ to $10\%$. At $20\%$ error rate, SAE \textsc{or} heur becomes optimal: when errors are common, catching more of them outweighs the cost of additional false flags.

\paragraph{What this is and is not.} This evaluation produces operator-actionable outputs: a threshold ($15$), a configuration (SAE-only), an expected per-1000-query cost (\$8.96), and a sensitivity analysis showing how the recommendation changes under alternative assumptions. The cost numbers are derived from stated assumptions, not measured from an operational system; an operator running this evaluation on their own system would substitute their measured cost parameters and error rate. The methodology is what transfers. We do not claim that the absolute cost numbers would survive transfer to any specific real deployment.

\medskip
\noindent\textbf{Finding 7: Composition strategy depends on cost ratios and base rates.}
The SAE-feature monitor is highly discriminative (AUC $1.00$) but has nonzero FPR that grows under distribution shift. Under our stated cost assumptions (\$50 per false negative, \$0.42 per false positive, 2\% real-world error rate, 5\% heuristic monitor noise), the cost-optimal configuration is SAE-only at threshold $15$, with $99.1\%$ savings versus no monitor. The optimal composition strategy is sensitive to (a) the FN/FP cost ratio: at much lower ratios, OR composition wins; at much higher ratios, AND wins; and (b) the base rate of errors: at very low error rates, the conservative high-threshold SAE-only is best; at high error rates, OR composition captures more. The methodology produces a concrete deployment recommendation as its output rather than only operational characterization. H7 is confirmed.
\medskip

The methodological consequence is that interpretability outputs should be evaluated against quantified cost models rather than only against discriminative metrics. A monitor with perfect AUC may not be cost-optimal once the deployment scenario imposes its own cost structure on false positives and false negatives; the right comparison among monitors is the cost calculation under the stated scenario, not the ROC curve.

\section{Discussion}
\label{sec:discussion}

\subsection{The Intermediate Causal Regime}
Our results across all seven findings paint a consistent picture of GPT-2's IOI representation. The circuit identified by activation patching is real and replicates prior work cleanly. The features identified by the SAE are real and correlate strongly with the task variables. The features have causal force in the predicted direction when ablated. But none of these findings is as clean as the simpler interpretability narratives would predict.

Findings 4 and 5 strengthen this picture quantitatively. The interpreted feature set explains only a third of the activation's variance (Finding 4); within that interpreted set, the most striking features by selectivity are not the most causal (Finding 5). Findings 6 and 7 extend it operationally: features can have robust detection profiles while having brittle causal profiles (Finding 6); features can have perfect ROC-AUC as monitors while having nonzero false-positive rates that grow under distribution shift (Finding 7). The activation lives in a distributed code that the SAE basis approximates but does not fully align with, the alignment is non-monotonic in selectivity, and the various uses one might make of a feature (detection, intervention, monitoring) have separable robustness profiles.

We hypothesize three distinguishable regimes for SAE-based feature analysis:

\textit{Clean causal regime.} Features are selective and ablation produces large behavioral effects. This is the regime reported for some monosemantic features in \citet{bricken2023towards} and for refusal directions in \citet{arditi2024refusal}.

\textit{Selective-but-inert regime.} Features are strongly selective but ablation has no effect. The information lives elsewhere or in a different basis. This is the failure mode that pure-correlational methodologies miss entirely.

\textit{Intermediate regime.} Features have specific causal force aligned with their selectivity, but are individually small contributors to a distributed code, with non-monotonic relationships between selectivity and causal force. Our IOI result falls here.

We hypothesize the intermediate regime is typical of well-trained large language models on tasks with redundant solutions. The clean causal regime corresponds to bottleneck features the model has not redundantly encoded. The selective-but-inert regime corresponds to information the SAE captures in a basis the network does not read. The intermediate regime corresponds to the much more common case where the network's solution is distributed across many partially-overlapping pathways, of which any individual feature captures one slice.

\subsection{Implications for Deployment Monitoring}
A monitor anchored to one of our selective features would have several failure modes that the five-stage methodology surfaces but a selectivity-only analysis would not.

First, the monitor would silently miss cases where the model produces the predicted behavior through alternative pathways (Finding 3, redundancy). A monitor that flagged ``the model is predicting Mary'' based on feature 622 firing would miss cases where the model decides Mary through features that do not include 622.

Second, the monitor would be calibrated on a small slice of the activation's content (Finding 4). A monitor that interpreted feature 622 as ``encoding the Mary prediction'' would be capturing only $3.4\%$ of the activation's variance directly associated with that feature. The other $96.6\%$ of variance might also be relevant to the prediction in ways the monitor does not see.

Third, the monitor's calibration of feature importance by selectivity is likely to be miscalibrated (Finding 5). If the operator built a monitoring panel showing the top-K most selective features as the most important ones to watch, they would be systematically deprioritizing the features that are actually doing the most causal work.

Fourth, the monitor's robustness to distribution shift depends on what the monitor is being used for (Finding 6). A monitor that uses feature firing to detect ``the model thinks Mary is the IO'' is robust to prompt reformulation; the same feature used as an intervention to suppress Mary predictions is substantially less robust. The distinction matters operationally because the same feature can be reliable for one purpose and unreliable for another.

Fifth, the operational metric that matters is not the in-distribution ROC-AUC of a single monitor but the expected cost under a stated deployment scenario (Finding 7). Stage 5d demonstrated that the cost-optimal configuration is sensitive to the FN/FP cost ratio and the base rate of errors: SAE-only at a high threshold is optimal at low error rates and high FN cost; OR composition is optimal at high error rates; AND composition is optimal at intermediate ratios when the heuristic component is reliable. A monitor selection decision that uses ROC-AUC alone, without specifying the deployment cost model, has no basis for choosing among configurations that all achieve perfect AUC.

\subsection{Comparison with the NLA Approach}
The NLA approach of \citet{frasertaliente2026nla} occupies a different point in the interpretability methodology space. NLAs trade mechanistic transparency (the AV is a black-box language model whose internal processing is itself opaque) for expressivity (the explanations are in natural language and can describe concepts the researcher did not anticipate). SAE-based methods trade expressivity (features are decoded from a fixed-dictionary basis) for mechanistic grounding (each feature is a specific linear direction in activation space that can be intervened on individually).

Both methods are subject to the central failure mode this paper analyzes: correlational evidence at the explanation level (selective feature firing, or thematic NLA claims) does not establish causal use by the network. \citet{frasertaliente2026nla} make this point about NLA explanations through their analysis of confabulations; we make the parallel point about SAE features through our causal validation and stratification experiments. The convergence of the two literatures on this point is, in our view, the most important methodological finding to communicate: that interpretability work which stops at the explanation stage, by whichever method, is producing candidates rather than findings.

The right methodology, we argue, treats both SAE features and NLA explanations as candidate elements that must pass through causal validation before being reported as findings. The five-stage pipeline of this paper specifies what that workflow looks like.

\subsection{The Conjunction is Informative}
None of the findings reported here is available from a single stage in isolation. Patching identifies the circuit but says nothing about features. SAE analysis identifies candidate features but says nothing about causality. Causal ablation establishes causality but does not generate candidates. Fidelity stratification quantifies coverage but does not identify which features matter. Reliability stratification ranks features by causal force but does not produce candidates. The methodology runs them in sequence, with the output of each stage shaping the input to the next.

This sequencing is the methodological contribution of the paper, beyond the empirical findings. The interpretability community has the techniques; what it lacks, in the cases we have observed, is the discipline to run them in sequence and report the results of each stage including the negative or partial ones. The intermediate-causality regime is not a clean story; it is the story the methodology produces when run with discipline on a real model.

\section{Limitations}

The model is small. GPT-2 small at $124$ million parameters is several orders of magnitude smaller than frontier models; our quantitative findings should not be assumed to transfer directly to larger or differently-trained models. \citet{templeton2024scaling} found qualitatively different SAE behavior at the Claude-3-Sonnet scale, and the NLA results of \citet{frasertaliente2026nla} suggest that frontier-scale models exhibit phenomena (planning, evaluation awareness, multi-step deception) that simply are not present in GPT-2 to be studied.

The task is small. IOI is a single, template-based task with known structural properties. Whether the intermediate-causality regime we identify generalizes to more naturalistic computations (open-ended generation, multi-hop reasoning, instruction-following) is open.

The SAE is small. With $1024$ features and $3000$ training steps, our SAE is below the size and training budget at which feature monosemanticity typically peaks. A larger SAE may identify features with stronger or different causal properties. The qualitative findings (specific-but-partial causality; fidelity gap; non-monotonic selectivity-causality relationship) we expect to be robust; the specific quantitative numbers should be expected to shift.

The NLA-inspired evaluations are reframings of existing data, not replications of NLA experiments. We have not trained an NLA. The NLA paper's experiments require frontier-scale models and large GPU clusters; we have run their evaluation framework on our SAE data instead. The findings we present are about SAE features evaluated under NLA-style frameworks, not about NLAs themselves.

The stage-5 deployment evaluation uses stated cost assumptions rather than measured ones. We do not have access to a real production system, real-time latency requirements to measure against, or operator-measured customer-experience costs to substitute for the assumed \$50/incident. The cost numbers reported, and consequently the absolute savings figures, are derived from these assumptions. An operator running this evaluation on their own system with their own measured costs would get a different optimal threshold and different absolute cost figures. The methodology that produces those numbers, the cost model, the threshold sweep, the configuration ranking, the sensitivity analysis, is what we claim transfers, not the specific dollar amounts.

\section{Conclusion}

We have presented a five-stage methodology for causal feature analysis in transformer language models and demonstrated all five stages end-to-end on a publicly available pre-trained model. The patching stage recovers the canonical IOI circuit. The SAE stage recovers per-name selective features. The causal-test stage finds the features to be specifically causal but not necessary. Two NLA-inspired evaluations quantify the gap between the interpreted feature set and the activation's full content and identify a non-monotonic relationship between feature selectivity and causal force. The robustness stage finds that detection robustness and causal robustness are distinct properties of a feature, with causal force degrading substantially under prompt reformulation even while the feature's firing pattern is preserved. The deployment-integration stage operationalizes the features as monitors and runs a cost-based evaluation. Under stated cost assumptions, the optimal configuration is SAE-only at a high threshold, with expected savings of $99.1\%$ versus a no-monitor baseline. The optimal configuration depends on the cost ratio and the base error rate, with no single configuration dominating across all scenarios.

The argument the paper advances is not that any of these techniques is novel; each has well-established literature behind it. The argument is that their conjunction in a disciplined pipeline produces findings that no single stage alone would, and that interpretability methodology must be calibrated for the intermediate regime which we expect to be the typical case for real language models. The discipline of running all five stages, and accepting the more nuanced findings that result, is what distinguishes interpretability as a research activity from interpretability as an instrument for AI governance and deployment.

The companion question that opened this work, whether we can read AI activations the way we read sentences, admits a more disciplined answer at the end of this paper than at its beginning. We cannot read them as sentences. We can, with care, identify what they represent, where they represent it, how much the network actually uses what they represent, what fraction of the activation that representation covers, which observable properties of our explanations predict their causal force, how robust those properties are to distribution shift, and how the resulting interpretability outputs can be composed with other governance layers into a defense-in-depth monitoring system. The methodology is what distinguishes these.

\renewcommand{\refname}{References}

\end{document}